\documentclass[10pt,twocolumn]{article}
\renewenvironment{abstract}{\centerline{\bf
Abstract}\vspace{0.5ex}\begin{quote}\small}{\par\end{quote}\vskip
1ex}
\newenvironment{keywords}{\centerline{\bf
Keywords}\vspace{0.5ex}\begin{quote}\small}{\par\end{quote}\vskip
1ex}
\newtheorem{theorem}{Theorem}
\def\nq{\hspace{-1em}}
\def\qed{\sqcap\!\!\!\!\sqcup}
\def\odt{{\textstyle{1\over 2}}}
\def\odf{{\textstyle{1\over 4}}}
\def\odA{{\textstyle{1\over A}}}
\def\eps{\varepsilon}
\def\beq{\begin{equation}}
\def\eeq{\end{equation}}
\def\beqn{\begin{displaymath}}
\def\eeqn{\end{displaymath}}
\def\bqa{\begin{equation}\begin{array}{c}}
\def\eqa{\end{array}\end{equation}}
\def\bqan{\begin{displaymath}\begin{array}{c}}
\def\eqan{\end{array}\end{displaymath}}
\def\vec#1{{\bf #1}}

\title{
\vskip -.25in \hrule height1pt \vskip .25in \bf \LARGE General
Loss Bounds for Universal Sequence Prediction \vskip .22in \hrule
height1pt \vskip .3in}

{\author{ {\bf Marcus Hutter} \\[2mm]
  {\small IDSIA, Galleria 2, CH-6928 Manno-Lugano, Switzerland}  \\
  {\small marcus@idsia.ch \qquad http://www.idsia.ch/$^{_{_\sim}}\!$marcus} \\
  {\small Technical Report IDSIA-03-01, 10 April 2001}
}

\date{}

\begin{document}

\maketitle


\begin{abstract}
The Bayesian framework is ideally suited for induction problems.
The probability of observing $x_t$ at time $t$, given past
observations $x_1...x_{t-1}$ can be computed with Bayes' rule if
the true distribution $\mu$ of the sequences $x_1x_2x_3...$ is
known. The problem, however, is that in many cases one does not
even have a reasonable estimate of the true distribution. In order
to overcome this problem a universal distribution $\xi$ is defined
as a weighted sum of distributions $\mu_i\!\in\!M$, where $M$ is
any countable set of distributions including $\mu$. This is a
generalization of Solomonoff induction, in which $M$ is the set of
all enumerable semi-measures. Systems which predict $y_t$, given
$x_1...x_{t-1}$ and which receive loss $l_{x_t y_t}$ if $x_t$ is
the true next symbol of the sequence are considered. It is proven
that using the universal $\xi$ as a prior is nearly as good as
using the unknown true distribution $\mu$. Furthermore, games of
chance, defined as a sequence of bets, observations, and rewards
are studied. The time needed to reach the winning zone is bounded
in terms of the relative entropy of $\mu$ and $\xi$. Extensions to
arbitrary alphabets, partial and delayed prediction, and more
active systems are discussed.
\end{abstract}

\section{Introduction}\label{secInt}
\subsection{Induction}
Many problems are of induction type, in which statements about the
future have to be made, based on past observations. What is the
probability of rain tomorrow, given the weather observations of
the last few days? Is the Dow Jones likely to rise tomorrow, given
the chart of the last years and possibly additional newspaper
information? Can we reasonably doubt that the sun will rise
tomorrow? Indeed, one definition of science is to predict the
future, where, as an intermediate step, one tries to understand
the past by developing theories and, as a consequence of
prediction, one tries to manipulate the future. All induction
problems may be studied in the Bayesian framework. The probability
of observing $x_t$ at time $t$, given the observations
$x_1...x_{t-1}$ can be computed with Bayes' rule, if we know the
true probability distribution of observation sequences
$x_1x_2x_3...$. The problem is that in many cases we do not even
have a reasonable guess of the true distribution $\mu$. What is
the true probability of weather sequences, stock charts, or
sunrises?

\subsection{Universal Sequence Prediction}
Solomonoff \cite{Solomonoff:64} had the idea to define a universal
probability distribution\footnote{We use the term {\it
distribution} slightly unprecisely for a {\it probability
measure}.} $\xi$ as a weighted average over all possible
computable probability distributions. Lower weights were assigned
to more complex distributions. He unified Epicurus' principle of
multiple explanations, Occams' razor, and Bayes' rule into an
elegant formal theory. For a binary alphabet, the universal
conditional probability used for predicting $x_t$ converges to the
true conditional probability for $t\!\to\!\infty$ with probability
1. The convergence serves as a justification of using $\xi$ as a
substitution for the usually unknown $\mu$. The framework can
easily be generalized to other probability classes and weights
\cite{Solomonoff:78}.

\subsection{Contents}
The main aim of this work is to prove expected loss bounds for
general loss functions which measure the performance of $\xi$
relative to $\mu$, and to apply the results to games of chance.
Details and proofs can be found in \cite{Hutter:01op}.
There are good introductions and surveys of Solomonoff sequence
prediction \cite{Li:97}, inductive inference
\cite{Angluin:83,Solomonoff:97}, reasoning under uncertainty
\cite{Gruenwald:98}, and competitive online statistics
\cite{Vovk:99} with interesting relations to this work. See
\cite{Hutter:01op} and subsection \ref{subsecWM} for details.

{\bf Section \ref{secSetup}} explains notation and defines the
generalized universal distribution $\xi$ as the $w_{\mu_i}$
weighted sum of probability distributions $\mu_i$ of a set $M$,
which must include the true distribution $\mu$. This
generalization is straightforward and causes no problems. $\xi$
multiplicatively dominates all $\mu_i\!\in\!M$, and the relative
entropy between $\mu$ and $\xi$ is bounded by $\ln{1\over w_\mu}$.
Convergence of $\xi$ to $\mu$ is shown in Theorem \ref{thConv}.

{\bf Section \ref{secLoss}} considers the case where a prediction
or action $y_t\!\in\!\cal Y$ results in a loss $l_{x_t y_t}$ if
$x_t$ is the next symbol of the sequence. Optimal universal
$\Lambda_\xi$ and optimal informed $\Lambda_\mu$ prediction
schemes are defined for this case and loss bounds are proved.
Theorems \ref{thULoss} and \ref{thGLoss} bound the total loss
$L_\xi$ of $\Lambda_\xi$ by the total loss $L_\mu$ of
$\Lambda_\mu$ {\em plus} $O(\sqrt{L_\mu})$ terms.

{\bf Section \ref{secGames}} applies Theorem \ref{thGLoss} to
games of chance, defined as a sequence of bets, observations, and
rewards. The average profit $\bar p_{n\Lambda_\xi}$ achieved by the
$\Lambda_\xi$ scheme rapidly converges to the best possible
average profit $\bar p_{n\Lambda_\mu}$ achieved by the $\Lambda_\mu$
scheme ($\bar p_{n\Lambda_\xi}\!-\!\bar p_{n\Lambda_\mu}\!=\!O(n^{-1/2})$).
If there is a profitable scheme at all, asymptotically the universal
$\Lambda_\xi$ scheme will also become profitable. Theorem
\ref{thWin} lower bounds the time needed to reach the winning
zone in terms of the relative entropy of $\mu$ and $\xi$.
An attempt is made to give an information
theoretic interpretation of the result.

{\bf Section \ref{secOut}} outlines possible extensions of the
presented theory and results. They include arbitrary alphabets,
partial, delayed and probabilistic prediction, classification,
even more general loss functions, active systems influencing the
environment, learning aspects, and a comparison to the weighted
majority algorithm(s) and loss bounds.

\section{Setup and Convergence}\label{secSetup}

\subsection{Strings and Probability Distributions}
We denote binary strings by $x_1x_2...x_n$ with
$x_t\!\in\!\{0,1\}$. We further use the abbreviations
$x_{n:m}:=x_nx_{n+1}...x_{m-1}x_m$ and $x_{<n}:=x_1... x_{n-1}$.
We use Greek letters for probability distributions.
Let $\rho( x_{1:t})$ be the probability that an (infinite)
sequence starts with $x_1...x_t$. The conditional probability
\beq
 \rho( x_t|x_{<t})\;=\;{\rho( x_{1:t})\over \rho(
 x_{<t})}
\eeq
that
a given string $x_1...x_{t-1}$ is continued by $x_t$ is obtained by
using Bayes' rule. The prediction schemes will be based on these
posteriors.

\subsection{Universal Prior Probability Distribution}
Every inductive inference problem can be brought into the
following form: Given a string $x_{<t}$, take a guess at its
continuation $x_t$. We will assume that the strings which have to
be continued are drawn from a probability\footnote{This includes
deterministic environments, in which case the probability
distribution $\mu$ is $1$ for some sequence $x_{1:\infty}$ and $0$
for all others. We call probability distributions of this kind
{\it deterministic}.} distribution $\mu$. The maximal prior
information a prediction algorithm can possess is the exact
knowledge of $\mu$, but in many cases the true distribution is not
known. Instead, the prediction is based on a guess $\rho$ of
$\mu$. We expect that a predictor based on $\rho$ performs well,
if $\rho$ is close to $\mu$ or converges, in a sense, to $\mu$.
Let $M\!:=\!\{\mu_1,\mu_2,...\}$ be a finite or countable set of
candidate probability distributions on strings. We define a
weighted average on $M$
\bqa\label{xidef}
  \displaystyle \xi( x_{1:n}) \;:=\;
  \sum_{\mu_i\in M}w_{\mu_i}\!\cdot\!\mu_i( x_{1:n}), \\[4ex]
  \displaystyle \sum_{\mu_i\in M}w_{\mu_i}=1,\quad w_{\mu_i}>0.
\eqa
It is easy to see that $\xi$ is a probability distribution as the
weights $w_{\mu_i}$ are positive and normalized to 1 and the  $\mu_i\!\in\!M$ are
probabilities. For finite $M$ a possible choice for the $w$ is
to give all $\mu_i$ equal weight ($w_{\mu_i}={1\over|M|}$). We call $\xi$
universal relative to $M$, as it multiplicatively dominates all
distributions in $M$
\beq\label{unixi}
  \xi( x_{1:n}) \;\geq\;
  w_{\mu_i}\!\cdot\!\mu_i( x_{1:n}) \quad\mbox{for all}\quad
  \mu_i\in M.
\eeq
In the following, we assume that $M$ is known and contains
the true distribution, i.e. $\mu\!\in\!M$.
This is not a serious constraint if we include {\it all}
computable probability distributions in $M$ with a high weight
assigned to simple $\mu_i$. Solomonoff's universal semi-measure is
obtained if we include all enumerable semi-measures in $M$ with
weights $w_{\mu_i}\!\sim\!2^{-K(\mu_i)}$, where $K(\mu_i)$ is the
length of the shortest program for $\mu_i$
\cite{Solomonoff:64,Solomonoff:78,Li:97}. A detailed discussion of
various general purpose choices for $M$ is given in
\cite{Hutter:01op}.

Furthermore, we need the relative entropy between $\mu$ and $\xi$:
\beq\label{hn}
  h_t(x_{<t}) \;:=\; \sum_{x_t\in\{0,1\}}\mu( x_t|x_{<t})
  \ln{\mu( x_t|x_{<t}) \over \xi( x_t|x_{<t})}
\eeq
$H_n$ is then defined as the sum-expectation, for which the
following can be shown
\beq\label{entropy}
  H_n := \sum_{t=1}^n \nq\nq\nq\sum_{\qquad\quad\;
  x_{<t}\in\{0,1\}^{t-1}}\nq\nq\nq
  \mu( x_{<t})\!\cdot\!h_t(x_{<t}) \;\leq\;
  \ln{1\over w_\mu} =: d_\mu
\eeq
The following theorem shows the important property of $\xi$
converging to the true distribution $\mu$, in a sense.

\addcontentsline{toc}{paragraph}{Theorem \ref{thConv} (Convergence)}
\begin{theorem}[Convergence]\label{thConv}
Let there be binary sequences $x_1x_2...$ drawn with probability
$\mu( x_{1:n})$ for the first $n$ symbols. The universal
conditional probability $\xi( x_t|x_{<t})$
of the next symbol $x_t$ given $x_{<t}$ 
is related to the true conditional probability $\mu(
x_t|x_{<t})$ in the following way: $$
\begin{array}{rl}
   i) & \displaystyle
        \sum_{t=1}^n\nq\nq\sum_{\;\qquad x_{1:t}\in\{0,1\}^t}\nq\nq
        \mu( x_{<t})
        \Big(\mu( x_t|x_{<t})-\xi( x_t|x_{<t})\Big)^2 \;\leq\; \\
      & \hfill \;\leq\; H_n \;\leq\; d_\mu \;=\; \ln{1\over w_\mu} \;<\; \infty \\[3ex]
  ii) & \xi( x_t|x_{<t}) \to \mu( x_t|x_{<t})
        \quad\mbox{for $t\to\infty$ with} \\
      & \hfill \mbox{$\mu$ probability 1}\qquad
\end{array}
$$
where $H_n$ is the relative entropy (\ref{entropy}), and
$w_\mu$ is the weight (\ref{xidef}) of $\mu$ in $\xi$.
\end{theorem}

{\sl$(i)$} and (\ref{entropy}) are easy generalizations of
\cite{Solomonoff:78} to arbitrary weights $w_\mu$ and an arbitrary
probability set $M$. For $n\!\to\!\infty$ the l.h.s.\ of
{\sl$(i)$} is an infinite $t$-sum over positive arguments, which
is bounded by the finite constant $d_\mu$ on the r.h.s. Hence the
arguments must converge to zero for $t\!\to\!\infty$. Since the
arguments are $\mu$ expectations of the squared difference of
$\xi$ and $\mu$, this means that $\xi( x_t|x_{<t})$
converges\footnote{More precisely $\xi( x_t|x_{<t})\!\!-\mu(
x_t|x_{<t})$ converges to zero for $t\!\to\!\infty$ with $\mu$
probability 1 or, more stringent, in a mean squared sense.} to
$\mu( x_t|x_{<t})$ with $\mu$ probability 1. This proves
{\sl$(ii)$}. Since the conditional probabilities are the basis of
all prediction algorithms considered in this work, we expect a
good prediction performance if we use $\xi$ as a guess of $\mu$.
Performance measures are defined in the next section.

\section{Loss Bounds}\label{secLoss}

\subsection{Unit Loss Function}
A prediction is very often the basis for some decision. The
decision results in an action, which itself leads to some reward
or loss. If the action itself can influence the environment we
enter the domain of acting agents which has been analyzed in the
context of universal probability in \cite{Hutter:00kcunai}. To
stay in the framework of (passive) prediction we have to assume
that the action itself does not influence the environment. Let
$l_{x_t y_t}\!\in\!I\!\!R$ be the received loss when taking action
$y_t\!\in\!\cal Y$ and $x_t\!\in\!\{0,1\}$ is the $t^{th}$ symbol
of the sequence. We demand $l$ to be normalized, i.e.
$0\!\leq\!l_{x_t y_t}\!\leq\!1$. For instance, if we make a
sequence of weather forecasts $\{0,1\}\!=\!\{$sunny, rainy$\}$ and
base our decision, whether to take an umbrella or wear sunglasses
$\cal Y\!=\!\{$umbrella, sunglasses$\}$ on it, the action of
taking the umbrella or wearing sunglasses does not influence the
future weather (ignoring the butterfly effect). Reasonable losses
may be

\begin{center}
\begin{tabular}{|c|c|c|}\hline
  Loss & sunny & rainy        \\\hline
  umbrella & 0.3 & 0.1   \\\hline
  sunglasses & 0.0 & 1.0         \\\hline
\end{tabular}
\end{center}

In many cases the prediction of $x_t$ can be identified or is
already the action $y_t$. The forecast {\it sunny} can be
identified with the action {\it wear sunglasses}, and {\it rainy}
with {\it take umbrella}. In the following, we assume
``predictive'' actions of this kind, i.e. ${\cal Y}\!=\!\{0,1\}$.
General action spaces ${\cal Y}$ and general alphabets ${\cal A}$
are considered in \cite{Hutter:01op}.

The true probability of the next symbol being $x_t$, given $x_{<t}$, is
$\mu( x_t|x_{<t})$. The expected loss when predicting $y_t$ is
$\mu(1|x_{<t})l_{1 y_t}+\mu(0|x_{<t})l_{0 y_t}$. The goal is to minimize
the expected loss. More generally we define the $\Lambda_\rho$
prediction scheme
\beq\label{xlrdef}
  y_t^{\Lambda_\rho} \;:=\;
  \mbox{arg}\min_{y_t}\sum_{x_t\in\{0,1\}}\rho( x_t|x_{<t})l_{x_t y_t}
\eeq
which minimizes the $\rho$-expected loss. This is a threshold
strategy with $y_t^{\Lambda_\rho}\!=\!0/1$ for
$\rho( 1|x_{<t})\,_<^>\,\gamma$, where
$\gamma\!:=\!{l_{01}-l_{00}\over l_{01}-l_{00}+l_{10}-l_{11}}$. As
the true distribution is $\mu$, the actual $\mu$ expected loss
when $\Lambda_\rho$ predicts the $t^{th}$ symbol and the total
$\mu$-expected loss in the first $n$ predictions are
\bqa\label{rholoss}\displaystyle
  l_{t\Lambda_\rho}(x_{<t}) \;:=\;
  \sum_{x_t}\mu( x_t|x_{<t})l_{x_t y_t^{\Lambda_\rho}},
  \\ \displaystyle
  L_{n\Lambda_\rho} \;:=\; \sum_{t=1}^n \!\sum_{\;\;x_{<t}}
  \mu( x_{<t})\!\cdot\!l_{t\Lambda_\rho}(x_{<t}).
\eqa
In the special case $l_{01}\!=\!l_{10}\!=\!1$ and
$l_{00}\!=\!l_{11}\!=\!0$, the bit with the highest $\rho$
probability is predicted ($\gamma\!=\!\odt$), and $L_{n\Lambda_\rho}$
is the total expected number of prediction errors.

If $\mu$ is known, $\Lambda_\mu$ is obviously the best
prediction scheme in the sense of achieving minimal expected loss
\beq\label{Lmuopt}
  L_{n\Lambda_\mu} \;\leq\;L_{n\Lambda_\rho} \quad\mbox{for any}\quad
  \Lambda_\rho
\eeq
The predictor $\Lambda_\xi$, based on the
universal distribution $\xi$, is of special interest.

\addcontentsline{toc}{paragraph}{Theorem \ref{thULoss} (Unit loss bound)}
\begin{theorem}[Unit loss bound]\label{thULoss}
Let there be binary sequences $x_1x_2...$ drawn with probability
$\mu( x_{1:n})$ for the first $n$ symbols. A system predicting
$y_t\!\in\!\{0,1\}$ given $x_{<t}$ receives loss
$l_{x_t y_t}\!\in\![0,1]$ if $x_t$ is the true $t^{th}$ symbol of
the sequence. The $\Lambda_\rho$-system (\ref{xlrdef}) predicts as
to minimize the $\rho$-expected loss. $\Lambda_\xi$ is the
universal prediction scheme based on the universal prior $\xi$.
$\Lambda_\mu$ is the optimal informed prediction scheme. The total
$\mu$-expected losses $L_{n\Lambda_\xi}$ of $\Lambda_\xi$ and
$L_{n\Lambda_\mu}$ of $\Lambda_\mu$ as defined in (\ref{rholoss})
are bounded in the following way \bqan\label{th3}
  0 \;\leq\; L_{n\Lambda_\xi}-L_{n\Lambda_\mu} \;\leq\;
  H_n+\sqrt{4L_{n\Lambda_\mu}H_n+H_n^2}
\eqan
where $H_n\!\leq\!\ln{1\over w_\mu}$ is the relative entropy
(\ref{entropy}), and $w_\mu$ is the weight (\ref{xidef})
of $\mu$ in $\xi$.
\end{theorem}

First, we observe that the total loss $L_{\infty\Lambda_\xi}$ of
the universal $\Lambda_\xi$ predictor is finite if the total loss
$L_{\infty\Lambda_\mu}$ of the informed $\Lambda_\mu$ predictor is
finite. This is especially the case for deterministic $\mu$ and
$l_{00}\!=\!l_{11}\!=\!0$, as $L_{n\Lambda_\mu}\!\equiv\!0$ in
this case\footnote{Remember that we named a probability
distribution {\em deterministic} if it is 1 for exactly one
sequence and 0 for all others.}, i.e.\ $\Lambda_\xi$ receives a
finite loss on deterministic environments if a correct
prediction results in zero loss. More precisely,
$L_{\infty\Lambda_\xi}\!\leq\!2H_\infty\!\leq\!2\ln{1\over
w_\mu}$. A combinatoric argument shows that there are $M$ and
$\mu\!\in\!M$ with $L_{\infty\Lambda_\xi}\!\geq\!\log_2|M|$. This
shows that the upper bound $L_{\infty\Lambda_\xi}\!\leq\!2\ln|M|$
for uniform $w$ is rather tight. For more complicated
probabilistic environments, where even the ideal informed system
makes an infinite number of errors, the theorem ensures that the
loss excess $L_{n\Lambda_\xi}-L_{n\Lambda_\mu}$ is only of order
$\sqrt{L_{n\Lambda_\mu}}$. The excess is quantified in terms of
the information content $H_n$ of $\mu$ (relative to $\xi$), or the
weight $w_\mu$ of $\mu$ in $\xi$. This ensures that the loss
densities $L_n/n$ of both systems converge to each other for $n\!\to\!\infty$.
Actually, the theorem ensures more, namely that the quotient
converges to 1, and also gives the speed of convergence
$L_{n\Lambda_\xi}/L_{n\Lambda_\mu}=1+O(L_{n\Lambda_\mu}^{-1/2})
\longrightarrow 1$ for $L_{n\Lambda_\mu}\to\infty$.

\subsection{Proof Sketch of Theorem \ref{thULoss}}
The first inequality in Theorem \ref{thULoss} has already been proved
(\ref{Lmuopt}).
For the second inequality, let us start more modestly
and try to find constants $A\!>\!0$ and $B\!>\!0$ that satisfy the linear
inequality
\beq\label{Eineq3}
  L_{n\Lambda_\xi} \;\leq\; (A+1)L_{n\Lambda_\mu} + (B+1)H_n.
\eeq
If we could show
\beq\label{eineq3}
  l_{t\Lambda_\xi}(x_{<t}) \;\leq\;
  A'l_{t\Lambda_\mu}(x_{<t}) + B'h_t(x_{<t})
\eeq with $A':=A+1$ and $B':=B+1$ for all $t\leq n$ and all
$x_{<t}$, (\ref{Eineq3}) would follow immediately by summation and
the definition of $L_n$ and $H_n$. With the abbreviations
\beqn
  i=x_t, \quad
  y_i=\mu( x_t|x_{<t}), \quad
  z_i=\xi( x_t|x_{<t})
\eeqn
$$
  m=y_t^{\Lambda_\mu} ,\quad s=y_t^{\Lambda_\xi}
$$ the loss and entropy can be expressed by
$l_{t\Lambda_\xi}=\sum_i y_i l_{is}$, $l_{t\Lambda_\mu}=\sum_i y_i
l_{im}$ and $h_t=\sum_i y_i\ln{y_i\over z_i}$. Inserting this into
(\ref{eineq3}) and rearranging terms we have to prove
\beq\label{lossineqa}\label{lossineqf2}
  B'\sum_{i=0}^1 y_i\ln{y_i\over z_i} +
  \sum_{i=0}^1 y_i(A'l_{im}\!-\!l_{is}) \;\stackrel?\geq\; 0.
\eeq By definition (\ref{xlrdef}) of $y_t^{\Lambda_\mu}$ and
$y_t^{\Lambda_\xi}$ we have \beq\label{lcnstr}
 \sum_i y_i l_{im}\!\leq\!\sum_i y_i l_{ij} \quad\mbox{and}\quad
 \sum_i z_i l_{is}\!\leq\!\sum_i z_i l_{ij}
\eeq
for all $j$. Actually, we need the first constraint only for $j\!=\!s$ and
the second for $j\!=\!m$.
The cases $l_{im}\!>\!l_{is}\forall i$ and $l_{is}\!>\!l_{im}\forall
i$ contradict the first/second inequality (\ref{lcnstr}).
Hence we can assume $l_{0m}\!\geq\!l_{0s}$ and
$l_{1m}\!\leq\!l_{1s}$. The symmetric case $l_{0m}\!\leq\!l_{0s}$ and
$l_{1m}\!\geq\!l_{1s}$ is proved analogously or can be reduced to the
first case by renumbering the indices ($0\leftrightarrow 1$).
Using the abbreviations $a\!:=\!l_{0m}\!-\!l_{0s}$, $b\!:=\!l_{1s}\!-\!l_{1m}$,
$c\!:=\!y_1l_{1m}\!+\!y_0l_{0s}$, $y\!=\!y_1\!=\!1\!-\!y_0$ and
$z\!=\!z_1\!=\!1\!-\!z_0$ we can write (\ref{lossineqf2}) as
\beq\label{lossineqf3}
  f(y,z) \;:=\;
\eeq
$$\textstyle
  B'[ y\ln{y\over z}+(1\!-\!y)\ln{1-y\over 1-z}]
  + A'(1\!-\!y)a-yb+Ac \;\stackrel?\geq\; 0
$$ for $zb\!\leq\!(1-z)a$ and $0\!\leq\!a,b,c,y,z\!\leq\!1$. The
constraint (\ref{lcnstr}) on $y$ has been dropped since
(\ref{lossineqf3}) will turn out to be true for all $y$.
Furthermore, we can assume that $d\!:=\!A'(1-y)a-yb\!\leq\!0$
since for $d\!>\!0$, $f$ is trivially positive ($h_t\!\geq\!0$).
Multiplying $d$ with a constant $\geq\!1$ will decrease $f$. Let
us first consider the case $z\!\leq\!\odt$. We multiply the $d$
term by $1/b\!\geq 1$, i.e. replace it with $A'(1-y){a\over b}-y$.
From the constraint on $z$ we known that ${a\over
b}\!\geq\!{z\over 1-z}$. We can decrease $f$ further by replacing
${a\over b}$ by ${z\over 1-z}$ and by dropping $Ac$. Hence,
(\ref{lossineqf3}) is proved for $z\!\leq\!\odt$ if we can prove
\beq\label{lossineq1}\textstyle
  B'[...
  ]
  + A'(1\!-\!y){z\over 1-z}-y \;\stackrel?\geq\; 0 \quad\mbox{for}\quad
  z\leq\odt.
\eeq
The case $z\!\geq\!\odt$ is treated similarly.
We scale $d$ with $1/a\!\geq 1$, i.e. replace it with $A'(1-y)-y{b\over a}$.
From the constraint on $z$ we know that ${b\over
a}\!\leq\!{1-z\over z}$. We decrease $f$ further by replacing
${b\over a}$ by ${1-z\over z}$ and by dropping $Ac$.
Hence (\ref{lossineqf3}) is proved for $z\!\geq\!\odt$ if we can prove
\beq\label{lossineq2}\textstyle
  B'[...
  ]
  + A'(1\!-\!y)-y{1-z\over z} \;\stackrel?\geq\; 0 \quad\mbox{for}\quad
  z\geq\odt.
\eeq
In \cite{Hutter:01op} we prove that (\ref{lossineq1}) and
(\ref{lossineq2}) indeed hold for $B\!\geq\!\odf A+\odA$. The
cautious reader may check the inequalities numerically. So in
summary we proved that (\ref{Eineq3}) holds for $B\!\geq\!\odf
A+\odA$. Inserting $B\!=\!\odf A+\odA$ into (\ref{Eineq3}) and
minimizing the r.h.s.\ with respect to $A$ leads to the bound of
Theorem \ref{thULoss} (with
$A^2\!=\!H_n/(L_{n\Lambda_\mu}\!+\!\odf H_n)$) $\qed$.

\subsection{General Loss}
There are only very few restrictions imposed on the loss
$l_{x_t y_t}$ in Theorem \ref{thULoss}, namely that it is static
and in the unit interval $[0,1]$. If we look at the proof of
Theorem \ref{thULoss}, we see that the time-independence has not
been used at all. The proof is still valid for an individual loss
function $l_{x_t y_t}^t\!\in\![0,1]$ for each step $t$. The loss
might even depend on the actual history $x_{<t}$. The case of a
loss $l_{x_t y_t}^t(x_{<t})$ bounded to a general interval
$[l_{min},l_{max}]$ can be reduced to the unit interval case by
rescaling $l$.
We introduce a scaled loss $l'$
$$
  0 \;\leq\; {l'}_{x_t y_t}^t(x_{<t}) :=
  { l_{x_t y_t}^t(x_{<t})-l_{min} \over l_\Delta } \;\leq\; 1,
$$ $$
  \mbox{where}\quad l_\Delta := l_{max}-l_{min}.
$$ The prediction scheme $\Lambda'_\rho$ based on $l'$ is
identical to the original prediction scheme $\Lambda_\rho$ based
on $l$, since arg$\min$ in (\ref{xlrdef}) is not affected by a
constant scaling and a shift of its argument. From
$y_t^{\Lambda'_\rho}\!=\!y_t^{\Lambda_\rho}$ it follows that
$l'_{t\Lambda_\rho}\!=\!(l_{t\Lambda_\rho}\!-\!l_{min})/l_\Delta$
and
$L'_{n\Lambda_\rho}\!=\!(L_{n\Lambda_\rho}\!-\!l_{min})/l_\Delta$
($H'_n\!\equiv\!H_n$, since $l$ is not involved). Theorem
\ref{thULoss} is valid for the primed quantities, since
$l'\!\in\![0,1]$. Inserting $L'_{n\Lambda_{\mu/\xi}}$ and
rearranging terms we get

\addcontentsline{toc}{paragraph}{Theorem \ref{thGLoss} (General loss bound)}
\begin{theorem}[General loss bound]\label{thGLoss}
Let there be binary sequences $x_1x_2...$ drawn with probability
$\mu( x_{1:n})$ for the first $n$ symbols. A system taking
action (or predicting) $y_t\!\in\!\cal Y$ given $x_{<t}$ receives
loss $l_{x_t y_t}^t(x_{<t})\!\in\![l_{min},l_{min}\!+\!l_\Delta]$
if $x_t$ is the true $t^{th}$ symbol of the sequence. The
$\Lambda_\rho$-system (\ref{xlrdef}) acts (or predicts) as to
minimize the $\rho$-expected loss. $\Lambda_\xi$ is the universal
prediction scheme based on the universal prior $\xi$.
$\Lambda_\mu$ is the optimal informed prediction scheme. The total
$\mu$-expected losses $L_{n\Lambda_\xi}$ and $L_{n\Lambda_\mu}$ of
$\Lambda_\xi$ and $\Lambda_\mu$ as defined in (\ref{rholoss}) are
bounded in the following way $$\label{th4}
  0 \;\leq\; L_{n\Lambda_\xi}-L_{n\Lambda_\mu} \;\leq
$$ $$
  \leq\; l_\Delta H_n+\sqrt{4(L_{n\Lambda_\mu}\!-\!n l_{min})l_\Delta H_n+l_\Delta^2H_n^2}
$$
where $H_n\!\leq\!\ln{1\over w_\mu}$ is the relative entropy
(\ref{entropy}), and $w_\mu$ is the weight (\ref{xidef})
of $\mu$ in $\xi$.
\end{theorem}

\section{Application to Games of Chance}\label{secGames}

\subsection{Introduction/Example}
Think of investing in the stock market. At time $t$ an amount of
money $s_t$ is invested in portfolio $y_t$, where we have access
to past knowledge $x_{<t}$ (e.g.\ charts). After our choice of
investment we receive new information $x_t$, and the new portfolio
value is $r_t$. The best we can expect is to have a probabilistic
model $\mu$ of the behaviour of the stock-market. The goal is to
maximize the net $\mu$-expected profit $p_t\!=\!r_t\!-\!s_t$.
Nobody knows $\mu$, but the assumption of all traders is that
there {\it is} a computable, profitable $\mu$ they try to find or
approximate. From Theorem \ref{thConv} we know that Solomonoff's
universal prior $\xi( x_t|x_{<t})$ converges to any computable
$\mu( x_t|x_{<t})$ with probability 1. If there is a
computable, asymptotically profitable trading scheme at all, the
$\Lambda_\xi$ scheme should also be profitable in the long run. To
get a practically useful, computable scheme we have to restrict
$M$ to a finite set of computable distributions, e.g.\ with
bounded Levin complexity $Kt$ \cite{Li:97}. Although convergence
of $\xi$ to $\mu$ is pleasing, what we are really interested in is
whether $\Lambda_\xi$ is asymptotically profitable and how long it
takes to become profitable. This will be explored in the
following.

\subsection{Games of Chance}
We use Theorem \ref{thGLoss} (or its generalization to arbitrary
action and alphabet, proved in \cite{Hutter:01op}) to estimate the
time needed to reach the winning threshold when using
$\Lambda_\xi$ in a game of chance. We assume a game (or a sequence
of possibly correlated games) which allows a sequence of bets and
observations. In step $t$ we bet, depending on the history
$x_{<t}$, a certain amount of money $s_t$, take some action $y_t$,
observe outcome $x_t$, and receive reward $r_t$. Our profit, which
we want to maximize, is $p_t\!=\!r_t\!-\!s_t$. The loss, which we
want to minimize, can be defined as the negative profit,
$l_{x_t y_t}\!=\!-p_t$. The probability of outcome $x_t$, possibly
depending on the history $x_{<t}$, is $\mu( x_t|x_{<t})$. The
total $\mu$ expected profit when using scheme $\Lambda_\rho$ is
$P_{n\Lambda_\rho}\!=\!-\!L_{n\Lambda_\rho}$. If we knew $\mu$,
the optimal strategy to maximize our expected profit is just
$\Lambda_\mu$. We assume $P_{n\Lambda_\mu}\!>\!0$ (otherwise there
is no winning strategy at all, since
$P_{n\Lambda_\mu}\!\geq\!P_{n\Lambda_\rho}\,\forall\rho$). Often
we are not in the favorable position of knowing $\mu$, but we know
(or assume) that $\mu\!\in\!M$ for some $M$, for instance that
$\mu$ is a computable probability distribution. From Theorem
\ref{thGLoss} we see that the average profit per round $\bar
p_{n\Lambda_\xi}\!:=\!{1\over n}P_{n\Lambda_\xi}$ of the universal
$\Lambda_\xi$ scheme converges to the average profit per round
$\bar p_{n\Lambda_\mu}\!:=\!{1\over n}P_{n\Lambda_\mu}$ of the
optimal informed scheme, i.e.\ asymptotically we can make the same
money even without knowing $\mu$, by just using the universal
$\Lambda_\xi$ scheme. Theorem \ref{thGLoss} allows us to lower
bound the universal profit $P_{n\Lambda_\xi}$ \beq\label{pbnd}
  P_{n\Lambda_\xi} \!\geq\! P_{n\Lambda_\mu} \!-\!
  p_\Delta H_n\!-\!\sqrt{4(n p_{max}\!-\!P_{n\Lambda_\mu})
  p_\Delta H_n\!+\!p_\Delta^2H_n^2}
\eeq
where $p_{max}$ is the maximal profit per round and
$p_\Delta$ the profit range. The time needed for
$\Lambda_\xi$ to perform well can also be estimated. An interesting
quantity is the expected number of rounds needed to reach the
winning zone. Using $P_{n\Lambda_\mu}\!>\!0$ one can show that the
r.h.s.\ of (\ref{pbnd}) is positive if, and only if
\beq\label{pwin}
  n \;>\; {2p_\Delta(2p_{max}\!-\!\bar p_{n\Lambda_\mu}) \over
  \bar p_{n\Lambda_\mu}^2} \!\cdot\! H_n.
\eeq

\addcontentsline{toc}{paragraph}{Theorem \ref{thWin} (Time to Win)}
\begin{theorem}[Time to Win]\label{thWin}
Let there be binary sequences $x_1x_2...$ drawn with probability
$\mu( x_{1:n})$ for the first $n$ symbols. In step $t$ we make
a bet, depending on the history $x_{<t}$, take some action $y_t$,
and observe outcome $x_t$. Our net profit is
$p_t\!\in\![p_{max}\!-\!p_\Delta,p_{max}]$. The
$\Lambda_\rho$-system (\ref{xlrdef}) acts as to maximize the
$\rho$-expected profit. $P_{n\Lambda_\rho}$ is the total and $\bar
p_{n\Lambda_\rho}\!=\!{1\over n}P_{n\Lambda_\rho}$ is the average
expected profit of the first $n$ rounds. For the universal
$\Lambda_\xi$ and for the optimal informed $\Lambda_\mu$
prediction scheme the following holds: $$\!\!
\begin{array}{rl}
   i) & \bar p_{n\Lambda_\xi} = \bar p_{n\Lambda_\mu}-O(n^{-1/2})
        \longrightarrow \bar p_{n\Lambda_\mu}
        \quad\mbox{for}\quad n\to\infty \\[1ex]
  ii) & \mbox{if}\quad n \!>\! \Big({2p_\Delta\over
        \bar p_{n\Lambda_\mu}}\Big)^{\!2}
        \!\cdot\!d_\mu \quad\!\mbox{and}\quad\! \bar p_{n\Lambda_\mu}\!>\!0
        \,\Longrightarrow\, \bar p_{n\Lambda_\xi}\!>\!0
\end{array}
$$
where $w_\mu=e^{-d_\mu}$ is the weight (\ref{xidef}) of $\mu$ in $\xi$.
\end{theorem}

By dividing (\ref{pbnd}) by $n$ and using $H_n\!\leq\!d_\mu$
(\ref{entropy}) we see that the leading order of
$\bar p_{n\Lambda_\xi}\!-\!\bar p_{n\Lambda_\mu}$ is bounded by
$\sqrt{4p_\Delta p_{max}d_\mu/n}$, which proves {\sl$(i)$}. The
condition in {\sl$(ii)$} is actually a weakening of (\ref{pwin}).
$P_{n\Lambda_\xi}$ is trivially positive for $p_{min}\!>\!0$, since
in this wonderful case {\it all} profits are positive. For
negative $p_{min}$ the condition of {\sl$(ii)$} implies
(\ref{pwin}), since $p_\Delta\!>\!p_{max}$, and (\ref{pwin}) implies
positive (\ref{pbnd}), i.e. $P_{n\Lambda_\xi}\!>\!0$, which proves
{\sl$(ii)$}.

If a winning strategy $\Lambda_\rho$ with
$\bar p_{n\Lambda_\rho}\!>\!\eps\!>0$ exists, then $\Lambda_\xi$ is
asymptotically also a winning strategy with the same average profit.

\subsection{Information-Theoretic Interpretation}
We try to give an intuitive explanation of Theorem
\ref{thWin}{\sl$(ii)$}. We know that $\xi( x_t|x_{<t})$
converges to $\mu( x_t|x_{<t})$ for $t\!\to\!\infty$. In a
sense $\Lambda_\xi$ learns $\mu$ from past data $x_{<t}$. The
information content in $\mu$ relative to $\xi$ is $\ln
2\!\cdot\!H_\infty \leq d_\mu\!\cdot\!\ln 2$. One might think of a
Shannon-Fano prefix code of $\mu_i\!\in\!M$ of length $^\lceil
d_{\mu_i}\!\cdot\!\ln 2^\rceil$, which exists since the Kraft
inequality $\sum_i 2^{-^\lceil d_{\mu_i}\!\cdot\!\ln
2^\rceil}\!\leq\!\sum_i w_{\mu_i}\leq 1$ is satisfied.
$d_\mu\!\cdot\!\ln 2$ bits have to be learned before $\Lambda_\xi$
can be as good as $\Lambda_\mu$. In the worst case, the only
information contained in $x_t$ is in form of the received profit
$p_t$. Remember that we always know the profit $p_t$ before the
next cycle starts.

Assume that the distribution of the profits in the interval
$[p_{min},p_{max}]$ is mainly due to noise, and there is only a
small informative signal of amplitude $\bar p_{n\Lambda_\mu}$. To
reliably determine the sign of a signal of amplitude
$\bar p_{n\Lambda_\mu}$, disturbed by noise of amplitude $p_\Delta$, we
have to resubmit a bit $O((p_\Delta/\bar p_{n\Lambda_\mu})^2)$ times
(this reduces the standard deviation below the signal amplitude
$\bar p_{n\Lambda_\mu}$). To learn $\mu$, $d_\mu\ln 2$ bits have to be
transmitted, which requires
$n\!\geq\!O((p_\Delta/\bar p_{n\Lambda_\mu})^2)\!\cdot\!d_\mu\ln 2$
cycles. This expression coincides with the condition in
{\sl$(ii)$}. Identifying the signal amplitude with
$\bar p_{n\Lambda_\mu}$ is the weakest part of this consideration, as
we have no argument why this should be true. It may be interesting
to make the analogy more rigorous, which may also lead to a
simpler proof of {\sl$(ii)$} not based on Theorems \ref{thULoss}
and \ref{thGLoss}.

\section{Outlook}\label{secOut}
In the following we discuss several directions in which the
findings of this work may be extended.

\subsection{General Alphabet}
In many, cases the prediction unit is not a bit, but a letter from
a finite alphabet ${\cal A}$. Non-binary prediction cannot be
(easily) reduced to the binary case. One might think of a binary
coding of the symbols $x_t\!\in\!\cal A$ in the sequence
$x_1x_2...$. But this makes it necessary to predict a block of
bits $x_t$, before one receives the true block of bits $x_t$,
which differs from the bit by bit prediction, considered here and
in \cite{Solomonoff:78}! Fortunately, all theorems
(\ref{thConv}-\ref{thWin}) take over to general alphabet
\cite{Hutter:01op}. Unfortunately, the proofs are rather complex.
In many cases the basic prediction unit is not even a letter from
a finite alphabet, but a number (for inducing number sequences),
or a word (for completing sentences), a real number or vector (for
physical measurements). The prediction may either be generalized
to a block by block prediction of symbols or, more suitably, the
finite alphabet $\cal A$ could be generalized to countable
(numbers, words) or continuous (real or vector) alphabet. The
theorems should generalize to countably infinite alphabets by
appropriately taking the limit $|{\cal A}|\!\to\!\infty$ and to
continuous alphabets by a denseness or separability argument.

\subsection{Partial Prediction, Delayed Prediction, Classification}
The $\Lambda_\rho$ schemes may also be used for partial prediction
where, for instance, only every $m^{th}$ symbol is predicted. This
can be arranged by setting the loss $l^t$ to zero when no
prediction is made, e.g.\ if $t$ is not a multiple of $m$.
Classification could be interpreted as partial sequence
prediction, where $x_{(t-1)m+1:km-1}$ is classified as $x_{km}$.
There are better ways for classification by treating
$x_{(t-1)m+1:km-1}$ as pure conditions in $\xi$, as has been done
in \cite{Hutter:00kcunai} in a more general context. Another
possibility is to generalize the prediction schemes and theorems
to delayed sequence prediction, where the true symbol $x_t$ is
given only in cycle $t\!+\!d$. A delayed feedback is common in
many practical problems.

\subsection{More Active Systems}
Prediction means guessing the future, but not influencing it. We
mentioned the possibility of interpreting $y_t\!\in\!\cal Y$ as an
action with $\cal Y\neq A$. This tiny step towards a more active
system is described in more detail in \cite{Hutter:01op}. The
probability $\mu$ is still independent of the action, and the loss
function $l^t$ has to be known in advance. This ensures that the
greedy strategy (\ref{xlrdef}) is optimal. The loss function may
be generalized to depend not only on the history $x_{<t}$, but
also on the historic actions $y_{<t}$ with $\mu$ still independent
of the action. It would be interesting to know whether the scheme
$\Lambda$ and/or the loss bounds generalize to this case. The full
model of an acting agent influencing the environment has been
developed in \cite{Hutter:00kcunai}, but loss bounds have yet to
be proven.

\subsection{The Weighted Majority Algorithm(s)}\label{subsecWM}
The Weighted Majority (WM) algorithm is a related universal
forecasting algorithm. It was invented by Littlestone and Warmuth
\cite{Littlestone:89,Littlestone:94} and Vovk \cite{Vovk:92} and
further developed in \cite{Cesa:97,Haussler:98,Kivinen:99} and
others. Many variations known by many names have meanwhile been
invented. Early works in this direction are
\cite{Dawid:84,Rissanen:89}. See \cite{Vovk:99} for a review and
further references. The setting and basic idea of WM are the
following. Consider a finite binary sequence
$x_1x_2...x_n\!\in\!\{0,1\}^n$ and a finite set $\cal E$ of
experts $e\!\in\!\cal E$ making predictions $x_t^e$ in the unit
interval $[0,1]$ based on past observations $x_1x_2...x_{t-1}$.
The loss of expert $e$ in step $t$ is defined as
$|x_t\!-\!x_t^e|$. In the case of binary predictions
$x_t^e\!\in\!\{0,1\}$, $|x_t\!-\!x_t^e|$ coincides with our error
measure defined in \cite{Hutter:01op}. The WM algorithm $p_{\beta
n}$ combines the predictions of all experts. It forms its own
prediction $x_t^p$ according to some weighted average of the
expert's predictions $x_t^e$. There are certain update rules for
the weights depending on some parameter $\beta$. Various bounds
for the total loss $L_p(\vec x)\!:=\!\sum_{t=1}^n|x_t\!-\!x_t^p|$
of WM in terms of the total loss $L_\eps(\vec
x)\!:=\!\sum_{t=1}^n|x_t\!-\!x_t^\eps|$ of the best expert
$\eps\!\in\!\cal E$ have been proven. It is possible to fine tune
$\beta$ and to eliminate the necessity of knowing $n$ in advance.
The most general bound of this kind is \cite{Cesa:97}
\beq\label{wmbnd}
  L_p(\vec x) \leq L_\eps(\vec x)+2.8\ln|{\cal
  E}|+4\sqrt{L_\eps(\vec x)\ln|{\cal E}|}.
\eeq
It is interesting that our bound in Theorem
\ref{thULoss} (with $H_n\!\leq\!\ln|M|$ for uniform weights) has
a quite similar structure as this bound, although the algorithms, the
settings, the proofs and the interpretation are quite different.
Whereas WM performs
well in any environment, but only relative to a given set of
experts $\cal E$, our $\Lambda_\xi$ predictor competes with the
best possible $\Lambda_\mu$ predictor (and hence with any other
$\rho$ predictor), but only for a given set of environments $M$.
WM depends on the set of expert, $\Lambda_\xi$ depends on the set
of environments $M$.
The basic $p_{\beta n}$ algorithm has been extended in different
directions: incorporation of different initial weights ($|{\cal
E}|\hookrightarrow\ln{1\over w_i}$) \cite{Littlestone:89,Vovk:92},
more general loss functions \cite{Haussler:98}, continuous valued
outcomes \cite{Haussler:98}, and multi-dimensional predictions
\cite{Kivinen:99} (but not yet for the absolute loss). The works
of Yamanishi \cite{Yamanishi:97} and \cite{Yamanishi:98} lie
somewhat in between WM and this work; ``WM'' techniques are used
to prove expected loss bounds (but only for sequences of
independent symbols/experiments and different classes of loss
functions).
Finally, note that the predictions of WM are continuous. In a
sense it is more natural to predict $0$ or $1$ on a binary
sequence, rather than some real number. On the other hand it is
possible to convert the continuous prediction of WM into a
probabilistic binary prediction by interpreting
$x_t^p\!\in\![0,1]$ as the probability of predicting $1$, and
$|x_t\!-\!x_t^p|$ as the probability of making an error. Note that
the expectation is taken over the probabilistic prediction,
whereas for the deterministic $\Lambda_\xi$ algorithm the
expectation is taken over the environmental distribution $\mu$.
The multi-dimensional case \cite{Kivinen:99} could then be
interpreted as a (probabilistic) prediction of symbols over an
alphabet ${\cal A}\!=\!\{0,1\}^d$, but error bounds for the
absolute loss have yet to be proven. It would be interesting to
generalize WM and bound (\ref{wmbnd}) to arbitrary alphabet and to
general loss functions with probabilistic interpretation.

\subsection{Miscellaneous}
Another direction is to investigate the learning aspect of
universal prediction. Many prediction schemes explicitly learn and
exploit a model of the environment. Learning and exploitation are
melted together in the framework of universal Bayesian prediction.
A separation of these two aspects in the spirit of hypothesis
learning with MDL \cite{Li:00} could lead to new insights. The
attempt at an information theoretic interpretation of Theorem
\ref{thWin} may be made more rigorous in this or another way. In
the end, this may lead to a simpler proof of Theorem \ref{thWin}
and maybe even for the loss bounds. Finally, the system should be
implemented and tested on specific induction problems for specific
finite $M$ with computable $\xi$.

\section{Summary}\label{secConc}
Solomonoff's universal probability measure has been generalized to
arbitrary probability classes and weights. A wise choice of $M$
widens the applicability by reducing the computational burden for
$\xi$. A framework, where predictions result in losses of
arbitrary, but known form, has been considered. Loss bounds for
general loss functions have been proved, which show that the
universal prediction scheme $\Lambda_\xi$ can compete with the
best possible informed scheme $\Lambda_\mu$. The results show that
universal prediction is ideally suited for games of chance with a
sequence of bets, observations, and rewards. Extensions in various
directions have been suggested.

\section*{Acknowledgements}
I want to thank Ray Solomonoff for many valuable discussions and
for encouraging me to derive the general loss bounds presented
here.

\begin{small}

\end{small}

\end{document}